\documentclass{article}

\usepackage{kr}

\usepackage{latexsym}
\usepackage{amssymb}
\usepackage{amsmath}
\usepackage{amsthm}
\usepackage{booktabs}
\usepackage{enumitem}
\usepackage{graphicx}
\usepackage{color}

\usepackage{algorithm}

\usepackage{times}
\usepackage{soul}
\usepackage{url}
\usepackage[hidelinks]{hyperref}
\usepackage[utf8]{inputenc}
\usepackage[small]{caption}
\usepackage{algorithmic}
\newcommand{\BibTeX}{B\kern-.05em{\sc i\kern-.025em b}\kern-.08em\TeX}

\usepackage{amsfonts}
\usepackage{etoolbox}
\usepackage{subfigure}

\usepackage{enumerate}
\usepackage{multirow}
\usepackage{tikz}
\usepackage{xcolor}
\usepackage{xspace}

\newcommand{\mlp}{\text{MLP}}

\newcommand{\comb}{\text{comb}}
\newcommand{\agg}{\text{agg}}
\newcommand{\multiset}[1]{\bigl\{\!\!\bigl\{#1\bigr\}\!\!\bigr\}}

\newcommand{\tup}[1]{\langle #1 \rangle}

\newcommand{\Omit}[1]{{}}
\newcommand{\bracket}[1]{[\![#1]\!]}

\newcommand{\Q}{\mathcal{Q}}

\DeclareMathOperator*{\argmin}{arg\,min}

\begin{document}

\title{Learning to Ground Existentially Quantified Goals}

\author{%
  Martin Funkquist$^1$\and
  Simon St\r{a}hlberg$^2$\and
  Hector Geffner$^{2}$ \\
  \affiliations
  $^1$Link\"{o}ping University, Sweden\\
  $^2$RWTH Aachen University, Germany\\
  \emails
  martin.funkquist@liu.se,
  simon.stahlberg@gmail.com,
  hector.geffner@ml.rwth-aachen.de
}

\maketitle

\begin{abstract}
  Goal instructions for autonomous AI agents cannot assume that objects have
unique names. Instead, objects in goals must be referred to by providing
suitable descriptions. However, this raises problems in
both classical planning and generalized planning. 
The standard approach to handling existentially quantified goals in classical 
planning involves compiling them into a DNF formula that encodes all possible 
variable bindings and adding dummy actions to map each DNF term into the new, dummy goal.
This preprocessing is exponential in the number of variables. In
generalized planning, the problem is different: even if general policies can
deal with any initial situation and goal, executing a general policy
requires the goal to be grounded to define a value for the policy features. The
problem of grounding goals, namely finding the objects to bind the goal
variables, is subtle: it is a generalization of classical planning, which
is a special case when there are no goal variables to bind, and constraint
reasoning, which is a special case when there are no actions. In this work, we
address the goal grounding problem with a novel supervised learning approach. 
A GNN architecture, trained to predict the cost of partially quantified
goals over small domain instances is tested on larger instances involving more
objects and different quantified goals. The proposed architecture is evaluated
experimentally over several planning domains where generalization
is tested along several dimensions including the number of goal variables and
objects that can bind such variables. The scope of the approach is
also discussed in light of the known relationship between GNNs and $C_2$ logics.

\end{abstract}

\section{Introduction}
In classical planning, the usual assumption is that objects have unique names,
and goals are described using these names. For instance, a goal might be to
place block A on top of block B, where A and B are specific blocks. A similar
assumption is made in generalized planning, where a policy is sought to handle
reactively any instances within a given domain. To apply the general plan to a
particular domain instance, it is assumed that the goal consists of a
conjunction of grounded atoms, with objects referred to by unique names.

However, instructions for goals in autonomous AI agents cannot assume that
objects have individual, known names. Instead, goals must be expressed by
referring to objects using suitable descriptions. For instance, the instruction
to place the large, yellow ball next to the blue package or to construct a tower
of 6 blocks, alternating between blue and red blocks, does not specify
the objects uniquely, and it's indeed part of the problem to select the
right objects. 

To address goals in classical planning that do not include unique object names,
existentially quantified goals are used~\cite{pednault-kr1989,gazen-knoblock-ecp1997,haslum-et-al-2019}. For
example, the goal of placing the large, yellow ball next to the blue
package can be expressed with the following formula:
\begin{align*}
    \exists x, y : \big[ & \textsc{Ball}(x) \wedge \textsc{Large}(x) \wedge \textsc{Yellow}(x)               \\
                         & \wedge \textsc{Package}(y)\wedge \textsc{Blue}(y) \wedge \textsc{Next}(x,y)\big].
\end{align*}

However, there are not many classical planners that support existentially
quantified goals because dealing with such goals is computationally hard, even
when actions are absent. Determining whether an existentially quantified goal is
true in a state is indeed NP-hard, as these goals can easily express constraint
satisfaction problems~\cite{frances-et-al-ijcai2016}. For example, consider a graph
coloring problem over a graph $G = \langle V, E \rangle$, where $V$ represents
vertices $i \in V$ and $E$ represents edges $(k,j) \in E$. This problem can be
expressed via the goal:
\[
    \exists x_1, \ldots, x_n : \Bigg[\bigwedge_{i \in V} \hspace{-0.25em}\textsc{Color}(i,x_i) \Bigg] \wedge \Bigg[\bigwedge_{(k,j) \in E} \hspace{-0.75em}\textsc{Neq}(x_k,x_j) \Bigg].
\]
This expression applies to an initial state with ground atoms
$\textsc{Color}(i,x)$, where $x$ is a possible color of vertex $i$, and
ground atoms $\textsc{Neq}(x,x')$ express that $x$ and $x'$ are
distinct.

The standard approach to handling existentially quantified goals in classical
planning involves eliminating them by transforming them into a grounded DNF
formula that encodes all possible variable bindings. This process also involves
adding dummy actions for each DNF term to map them to a new, dummy goal
\cite{gazen-knoblock-ecp1997}. However, this preprocessing step is exponential in the
number of variables. For instance, in the graph coloring example, it results in
several terms and dummy actions that grow exponentially with the number of
vertices $n$.

The problem of grounding goals, which involves finding objects to bind the goal
variables, is subtle. It is a generalization of standard classical
planning where there are no goal variables to bind, and it is a special case of
constraint reasoning, which is computationally hard even without considering
actions. Conceptually, the interesting problems lie between these two extremes,
in the presence of multiple possible bindings for the goal variables, each resulting
in a fully grounded goal with a "cost" determined by the number of steps needed to
reach it from the initial state. \emph{Optimal bindings}, or \emph{groundings},
replace the goal variables with constants to achieve grounded goals with minimum
cost. In action-less problems representing constraint satisfaction problems, the
cost of the bindings (i.e., the cost of the resulting grounded goals) is either
zero or infinity. In planning problems with existentially quantified goals, the
cost of the possible groundings depends on the initial state, the actions, and
the structure of the goal, and can be any natural number or infinity. For
example, in the planning problem with the initial and goal states shown in
Figures~\ref{fig:blocks-example}~and~\ref{fig:blocks-goal-example} below,
the optimal grounding binds the two bottom blocks in the goal to F and A, so that the first
move places A on F.

In this work, we tackle the problem of grounding goals as a \emph{generalized
planning
problem}~\cite{khardon-aij1999,martin-geffner-ai2004,fern-et-al-jair2006}. Here,
both the training and test instances are assumed to contain partially quantified
goals; that is, a combination of grounded and existentially quantified
variables. We use a graph neural network (GNN) architecture, trained using
supervised learning to predict the cost $V(P)$ of partially quantified goals
over small domain instances $P$ and test it on larger instances $P'$ involving
more objects and different goals. The bindings of the goal variables
are then obtained sequentially by greedily grounding one variable at a
time, without any search, neither in the problem state space nor the space of
goal bindings. Once the variables in the goal of an instance $P$ have been
grounded, %
existing methods can be used to obtain a plan for  the
fully grounded problem $P'$, %
using  standard  classical or  generalized planners
that expect grounded goals.

The paper is organized as follows: we start by explaining the learning task
through an example, and then review classical planning, generalized planning,
and existentially quantified goals. Next, we describe the %
task, introduce our proposed learning method,  present the experimental
results and detailed analysis over an example.

\section{Example: Visit-1}
\label{sec:example}

We begin with an example to illustrate our learning task, estimating the cost of
existentially quantified goals in planning, and why GNNs provide a handle
on this problem given the known correspondence between GNNs and $C_2$, the fragment of
first-order logic with two variables and counting \cite{barcelo-et-al-iclr2020,grohe-lics2021}.
Indeed, we show that the general value (cost) function for the problem can be expressed in $C_2$
and hence can be learned with GNNs. 

The example is a variation of the Visitall domain %
where a robot is placed on a %
grid and can move up, down, left, or right.
In the original domain, the goal is for the robot to visit all the cells in the grid.
Each cell is represented as a single object, and two cells $x$ and $y$ are considered adjacent if there is a $\textsc{Connected}(x, y)$ atom in the state.
A cell $x$ is marked as visited with a $\textsc{Visited}(x)$ atom.
The location of the robot is given by $\textsc{At-Robot}(x)$, where $x$ is a cell.

In our variation, cells have colors from a fixed set of colors $\mathcal{C}$
and the goal is to visit a cell of a given color. The cost of the problem
is thus given by the distance to the closest cell of that color.
For reasons to be elaborated later, the distances involved
must be bounded with the set $\mathcal{D}$ representing
the distances up to such a bound. 
The goals have the form: 
\[
    G = \exists x : \textsc{C}(x) \wedge \textsc{Visited}(x), 
    \]
\noindent  \text{where } C is any of the colors in $\mathcal{C}$.
Later on, we will consider a further variation of the problem where cells of different colors
are to be visited. The objective is to express the cost function for this family
of problems using Boolean features definable in $C_2$ so that the cost function
can be learned with GNNs. 

The Boolean function $\text{SP}_{d, \textsc{C}}(x)$ determines the existence of a shortest path of length $d$ from $x$ to a cell with color $\textsc{C}$:
\begin{align*}
    \text{P}_{0, \textsc{C}}(x) &= \textsc{C}(x) \\
    \text{P}_{d, \textsc{C}}(x) &= \exists y : \textsc{Connected}(x, y) \wedge \text{P}_{d - 1, \textsc{C}}(y) \\
    \text{SP}_{d, \textsc{C}}(x) &= \text{P}_{d, \textsc{C}}(x) \wedge \neg \text{P}_{d - 1, \textsc{C}}(x)
\end{align*}

If we let $N$ stand for the max distance plus 1, the value function $V^*$ can then be
expressed as:
\begin{align*}
    \textsc{G}_\textsc{C} &= \exists v : \textsc{C}_G(v) \\
    D_{d, \textsc{C}} &= \textsc{G}_\textsc{C} \wedge \exists x : \left[ \textsc{At-Robot}(x) \wedge \text{SP}_{d, \textsc{C}}(x) \right] \\
    V^*(s; G) &= \min_{d \in \mathcal{D}, \textsc{C} \in \mathcal{C}} d \cdot \bracket{D_{d, \textsc{C}}} + N \cdot \bracket{\neg D_{d, \textsc{C}}}
\end{align*}
Here, $s$ represents a state, and $G$ denotes a goal.
The notation $\bracket{\cdot}$ is the Iverson bracket, and $\textsc{C}_G$ refers to the colors specified in $G$.
When predicate names are used without subscripts, they refer to $s$.
If we assume that the state $s$ does not contain any $\textsc{Visited}$ atoms, then $V^*$ calculates the distance to the nearest cell with the color $\textsc{C}$, provided such a cell exists.
If no such cell exists, the value is set to $N$.

We will learn $V^*$ using GNNs. Since the expressiveness of GNNs is limited by two-variable first-order logic $C_2$,
if $V^*$ cannot be expressed with $C_2$ features, it will not be learnable by GNNs. Existentially quantified goals
are thus handled ``by free'' by GNNs as long as they do not get out of these limits. Moreover, existentially
quantified goals with more than two variables are not necessarily a problem if they are equivalent to formulas
in $C_2$. For example, the goal of building a tower of five blocks is naturally written in terms of five variables,
yet the formula is equivalent to a formula with two variables only that are quantified multiple times. 

\Omit{
It is important to note that the colors in $\mathcal{C}$ are used as subscripts in some functions.
Given a fixed number of distances and colors, the number of Boolean functions is also limited to a constant.
The number of layers in our architecture, described in Section~\ref{sec:architecture}, restricts the distances we can calculate.
As a result, each combination of distance and color can be represented as a Boolean feature in the embeddings.
This means that $V^*$ can be expressed using $C_2$, and our architecture might be able to learn it.
Additionally, since the minimization involves a fixed number of arguments, it can be approximated using an MLP.
}

\section{Related work}

\noindent \textbf{Quantification in planning.} %
Since the middle nineties, planners usually ground all actions and goals to
improve efficiency. This does not rule out the use of existential and universal
quantification in action preconditions and effects, and goals
\cite{pednault-kr1989,haslum-et-al-2019}. Universal quantification can be
replaced by conjunctions, while existential quantification by
disjunctions~\cite{gazen-knoblock-ecp1997}. The problem with existential
quantification in the goal %
is that the number of resulting disjuncts
is exponential in the number of goal variables. In action preconditions, the
problem is less critical as the number of variables is bounded by the arity of
the action schemas.

\medskip \noindent \textbf{Lifted planning.} %
Modern, lifted planners aim to approximate the performance of fully grounded
planners without having to ground actions or
goals~\cite{correa-et-al-icaps2020,stahlberg-ecai2023}. The problem with
non-grounded goals is that they do not result in equally informed
heuristics, and approaches that aim to compute informed heuristics without
grounding the actions or goals carry a significant
overhead~\cite{frances-et-al-ijcai2016}.

\medskip \noindent \textbf{Generalized planning.} %
The problem of learning policies for solving collections of problems involving
different number of objects and goals have been approached with
symbolic~\cite{khardon-aij1999,martin-geffner-ai2004,fern-et-al-jair2006,srivastava-et-al-aij2011,illanes-mcilraith-aaai2019,frances-et-al-aaai2021}
and deep learning
methods~\cite{toyer-et-al-jair2020,bajpai-et-al-neurips2018,rivlin-et-al-icaps2020wsprl,stahlberg-et-al-kr2022,stahlberg-et-al-kr2023},
yet in practically all cases, goals are assumed to be fully grounded.

\medskip \noindent \textbf{Grounding instructions in RL.} %
In reinforcement learning, the problem of grounding instructions is the problem
of understanding and carrying out the given
instructions~\cite{chevalier-boisvert-et-al-iclr2019,jaakola-et-al-jair2018,ruis-et-at-neurips2020}.
The key difference with generalized planning is that the structure of the states and goals, 
both sets of ground atoms over a fixed set of predicates in
planning, are not assumed to be known in RL. Instead, if a state or trajectory
is produced that complies with instructions, a reward is obtained. 

\medskip \noindent \textbf{GNNs and $C_k$ logics.} %
There is a tight correspondence among the classes of graphs that can be
distinguished by GNNs, the Weisfeiler-Leman algorithm
(1-WL)~\cite{morris-et-al-aaai2019,xu-et-al-iclr2019}, and two-variable
first-order logic with counting quantifiers
($C_2$)~\cite{cai-furer-immerman-combinatorica1992,barcelo-et-al-iclr2020,grohe-lics2021}.
Briefly, this means that $C_2$ serves as an upper bound of expressivity for
GNNs.

\section{Background}

We review classical planning, generalized planning, and existentially quantified goals.

\subsection{Classical Planning}

A classical planning problem is a pair $P\,{=}\,\tup{D,I}$ where $D$ is a first-order
\emph{domain} and $I$ contains information about the instance~\cite{geffner-bonet-2013,ghallab-et-al-2016,haslum-et-al-2019}.
The domain $D$ has a set of predicate symbols $p$ and a set of action schemas with
preconditions and effects given by atoms $p(x_1, \ldots, x_k)$ where $p$ is a predicate
symbol of arity $k$, and each $x_i$ is an argument of the schema.
An instance is a tuple $I\,{=}\,\tup{O, \textit{Init},\textit{Goal}}$ where $O$ is a
set of object names $c_i$, and $\textit{Init}$ and $\textit{Goal}$ are sets of
\emph{ground atoms} $p(c_1, \ldots, c_k)$.

A %
problem $P\,{=}\,\tup{D,I}$ encodes a state model
$S(P)\,{=}\,\tup{S,s_0,S_G,\textit{Act},A,f}$ in compact form where the states
$s \in S$ are sets of ground atoms, $s_0$ is the initial state $I$, $S_G$ is the
set of goal states $s$ such that $S_G \subseteq s$, $\textit{Act}$ is the set of
ground actions, $A(s)$ is the set of ground actions whose preconditions are true
in $s$, and $f$ is the induced transition function where $s'\,{=}\,f(a,s)$ is
the resulting state after applying $a \in A(s)$ in state $s$. An action sequence
$a_0, \ldots, a_{n}$ is applicable in $P$ if $a_i \in A(s_i)$ and
$s_{i+1}\,{=}\,f(a_i,s_i)$, for $i\,{=}\,1,\ldots,n$, and it is a plan if
$s_{n+1} \in S_G$.

The \emph{cost of a plan} is assumed to be given by its length and a plan is
\emph{optimal} if there is no shorter plan. The \emph{cost of a goal} $G$ for a
problem $P$ is the cost of the optimal plan to reach $G$ from the initial state of $P$.

\subsection{Generalized Planning}

In generalized planning, one is interested in solutions to collections $\Q$ of
problems $P$ over the same planning
domain~\cite{khardon-aij1999,martin-geffner-ai2004,fern-et-al-jair2006}. For
example, the class of problems $\Q$ may include all Blocksworld instances where
a given block $x$ must be cleared, or all instances of Blocksworld for any
(grounded) goal. A critical issue in generalized planning is the representation
of these general solutions or policies $\pi$ which must select one or more
actions in the reachable states $s$ of the instances $P \in \Q$. A %
common representation of these policies is in terms of general value functions
$V$~\cite{frances-et-al-ijcai2019,stahlberg-et-al-icaps2022} that map states $s$
into non-negative scalar values $V(s)$. The general policy $\pi_V$ greedy on $V$
then selects the action $a$ applicable in $s$ that result in successor state
$s'$ with minimum $V(s')$ value. If the value of the child $s'$ is always lower
than the value of its parent state $s$, the value function $V$ represents a
general policy $\pi_V$ that is guaranteed to solve any problem in the class
$\Q$.

In this formulation, it is assumed that the state $s$ over a problem $P$ in $\Q$
also encodes the goal $G$ of $P$ given by a set of ground atoms $p_G(c_1,
\ldots, c_k)$ for each ground goal $p(c_1, \ldots, c_k)$ in $P$. The new goal
predicate $p_G$ \cite{martin-geffner-ai2004} is used to indicate in the state $s$ that
the atom $p(c_1, \ldots, c_k)$ is to be achieved from $s$ and that it is not
necessarily true in $s$.

\subsection{Existentially Quantified Goals}

A classical planning problem with partially quantified goals is a pair
$P_X\,{=}\,\tup{D,I_X}$ where $D$ is a first-order \emph{domain}, $X$ is a set
of variables, and $I_X {=} \tup{O, \textit{Init},\textit{G}_X}$ expresses the
instance information as before, with one difference: the atoms $p(t_1, \ldots,
t_k)$ in the goal $G_X$ can contain \emph{variables} $x$ from $X$, which are
assumed to be existentially quantified. That is, in quantified problem $P_X$,
the terms $t_i$ can be either constants $c_i$ from $O$ referring to the objects
or variables $x$ from $X$. The variables $x$ in the goal $G_X$ of $P_X$
introduce a small change in the semantics of a standard classical planning
problem, where a state $s$ over $P_X$ is a \emph{goal state} if there is a
substitution of the variables $x_i$ in $G_X$ by constants $c_i$, $x_i \mapsto
c_i$, $x_i \in X$, $c_i \in O$, such that the resulting fully grounded goal
$G_C$ is true in $s$; namely, $G_C \subseteq s$. Quantification is sometimes used in 
action preconditions and sometimes involves universal
quantification, but we will leave this to future work.

As mentioned above, existential quantification in goals adds a second source of
complexity in planning, as even in the absence of actions, planning with
existentially quantified goals is NP-hard ~\cite{frances-et-al-ijcai2016}.
Most existing classical planners do not support existentially quantified goals, and
those that do, compile the goal variables away by considering all the possible
goal groundings and new, dummy actions, that map each one of them into a new
dummy goal that is to be reached. The problem with this approach is that it is
exponential in the number of goal variables. Lifted planning approaches can deal
with quantified preconditions and goals without having to ground
them~\cite{correa-et-al-icaps2020,stahlberg-ecai2023}, yet variables in the goal
affect the quality of the heuristics that can be obtained, and approaches that
aim to obtain more informed heuristics have a costly overhead
\cite{frances-et-al-icaps2015,frances-et-al-ijcai2016}

\section{Task: Learning to ground goals}

The problem of generalized planning with partially (existentially) quantified
goals can be split into two; namely, learning to ground the goals; i.e.,
substituting the goal variables with constants, and learning a general policy for
fully grounded goals. Since the second problem has been addressed in the
literature, we focus solely on the first part: learning to ground a given
partially quantified goal in a planning problem $P$ that belongs to a large
class of instances $\Q$ over the same planning domain but which may differ from
$P$ on several dimensions including the number of objects and the number of
goal atoms or variables.

We approach this learning task in a simple manner: by learning to predict the
optimal cost of partially grounded goals $G$ in families of problem instances
$P$ from a given domain $D$. Recall that the cost of $G$ is the cost of $P$ if
$G$ is the goal of $P$. %

For making these cost predictions, a general value function $V$ is learned
to approximate the optimal cost function $V^*$. The value function $V$
accepts a state $s$ and a partially quantified goal $G$ and outputs a
non-negative scalar that estimates the (min) number of steps to reach a state
$s'$ from $s$ in $P$ such that $s'$ satisfies the goal $G$ (i.e., there is a
grounding $G'$ of $G$ that is true in $s'$). We write the target value function
as $V(s;G)$ when we want to make explicit the goal $G$, else we write it simply
as $V(s)$.

We learn the target value function $V(s;G)$ over a given domain, where $G$ is a
partially quantified goal, in a supervised manner. Namely, for several
small instances $P$ from the domain, we use as targets for $V(s;G)$, the
optimal cost $V^*(s;G)$ of the problems $P[s,G]$ that are like $P$ but with
initial state $s$ and goal $G$. The learned value function $V(s;G)$ is expected
to generalize among several dimensions: different initial states, instances with more objects, 
and different goals with more goal atoms, variables, or both.

The learned value function $V(s;G)$ can then be used to bind the variables in
the partially quantified goal. %
For binding a single variable in $G$, %
we consider the goals $G'=G_{x=c}$
that result from $G$ by instantiating each variable $x$ in $G$ to a constant
$c$, while greedily choosing the goal $G'$ that minimizes
$V(s;G')$. For binding all the variables in $G$, the process is repeated until
a fully grounded goal is obtained. 

The quality of these goal groundings can then be determined by the ratio
$V^*(s;G')/V^*(s;G)$ where $V^*(s;G')$ is the optimal cost of achieving the
grounded goal $G'$ and $V^*(s;G)$ is the optimal cost of achieving the partially
quantified goal $G$. This ratio is $1$ when the grounded goal $G'$ is optimal
and else is strictly higher than $1$. For large instances, for which the optimal
values cannot be computed, $V^*$ values are replaced by $V^L$ values obtained
using a non-optimal planner that accepts existentially quantified goals.

The ability to map quantified goals $G$ into fully grounded goals $G$' can be
used in two different ways. In classical planning, it can be used to seek plans
for the quantified goals by seeking plans for the fully grounded goal $G'$,
while in generalized planning, it can be used to apply a learned general policy
$\pi$ for achieving a partially quantified goal $G$: for this $G$ is replaced
by $G'$.

\section{Architecture}
\label{sec:architecture}

We use Graph Neural Networks (GNNs) to learn how to bind variables to constants.
Since plain GNNs can only process graphs and not relational structures, we use
a suitable variant~\cite{stahlberg-et-al-icaps2022}. We describe GNNs first and then this extension.

\subsection{Graph Neural Networks}

GNNs are parametric functions that operate on graphs through aggregate and combination functions, denoted $\agg{}_i$ and $\comb{}_i$, respectively~\cite{scarselli-et-al-ieeenn2009,gilmer-et-al-icml2017,hamilton-2020}.
GNNs maintain and update embeddings $f_i(v) \in \mathbb{R}^k$ for each vertex $v$ in a graph $G$.
This process is performed iteratively over $L$ layers, starting with $i=0$ and the initial embeddings $f_0(v)$, and progressing to $f_{i+1}(v)$ as follows:
\begin{equation}
    f_{i+1}(v) = \comb{}_i\bigl(f_i(v), \agg{}_i\bigl(\multiset{f_i(w) \mid w \in N_G(v)}\bigr)\bigr)
    \label{eq:gnn:update}
\end{equation}
where $N_G(v)$ is the set of neighboring nodes of $v$ in the graph $G$.
The aggregation function $\agg{}_i$ (e.g., max, sum, or smooth-max) condenses multiple vectors into a single vector, while the combination function $\comb{}_i$ merges pairs of vectors.
The function implemented by GNNs is well-defined for graphs of any size and is invariant under graph isomorphisms, provided that the aggregation functions are permutation-invariant.

\subsection{Relational GNNs}

GNNs operate over graphs, whereas planning states $s$ are relational structures based on predicates of varying arities.
Our relational GNN (R-GNN) for processing these structures is inspired by the approach used for solving  min-CSPs~\cite{toenhoff-et-al-ewsp2021}
and closely follows the one used for learning general policies~\cite{stahlberg-et-al-icaps2022}, where the objects $o_i$ in a relational structure
(state) exchange messages with the objects $o_j$ through the atoms $q=p(o_1,\ldots,o_m)$ in the structure that involves the two objects and possibly others.
For dealing with existentially quantified variables $x$, the variables $x$ in the goal are treated as extra objects in the state.
The R-GNN shown in Algorithm~\ref{alg:relnn}, maps a state $s$ into a final embedding $f_L(o)$ for each object in the state (including the variables),
which feed a readout function and outputs the value $V(s)$ to be learned by adjusting the weights of the network.

In the R-GNN, there are atoms instead of edges, and messages are passed among objects appearing in the same atom:
\begin{equation*}
    f_{i+1}(o) = \comb{}_i\bigl(f_i(o), \agg{}_i\bigl(\multiset{m_{q,o} \mid o \in q, q \in S}\bigr)\bigr) \,.
\end{equation*}
Here, $m_{q,o}$ for a predicate $q = p(o_1, \ldots, o_m)$ and object $o = o_j$ represents the message that atom $q$ sends to object $o$, defined as:
\begin{displaymath}
    m_{q,o_j} = \bigl[\comb{}_p\bigl(f_i(o_1), \ldots, f_i(o_m)\bigr)\bigr]_j \,,
\end{displaymath}
where $\comb{}_p$ is the combination function for predicate $p$, generating $m$ messages, one for each object $o_j$, from their embeddings $f_i(o_j)$.
The $\comb{}_i$ function merges two vectors of size $k$, the current embedding $f_i(o)$ and the aggregation of the messages $m_{q,o}$ received at $o$.

\begin{algorithm}[t]
    \begin{algorithmic}[1]
        \STATE \textbf{Input:} Set of atoms $\mathcal{A}$, and set of objects $\mathcal{O}$
        \STATE \textbf{Output:} Embeddings $f_L(o)$ for each $o \in \mathcal{O}$
        \STATE Initialize $f_0(o) \sim 0^k$ for each node $o \in \mathcal{N}$
        \FOR{$i \in \{0, \ldots, L-1\}$}
        \FOR{each atom $q := p(o_1, \ldots, o_m) \in \mathcal{A}$}
        \STATE $m_{q,o_j} := [\mlp{}_p(f_i(o_1), \ldots, f_i(o_m))]_j$ %
        \ENDFOR
        \FOR{each object $n \in \mathcal{O}$}
        \STATE $f_{i+1}(n) := \mlp{}_U\bigl(f_i(o), \agg{}\bigl(\multiset{m_{q,o} \mid o \in q}\bigr)\bigr)$
        \ENDFOR
        \ENDFOR
    \end{algorithmic}
    \caption{%
    The Relational GNN (R-GNN) architecture used to predict partially quantified goals.
    }
    \label{alg:relnn}
\end{algorithm}

In our implementation, all layers share weights, and the aggregation function $\agg$ is the \emph{smooth maximum}, which approximates the component-wise maximum.
Each $\mlp$ consists of three parts: first, a linear layer; next, the Mish activation function~\cite{misra-bmva2020}; and then another linear layer.
The functions $\comb{}_i$ and $\comb{}_p$ correspond to $\mlp{}_U$ and $\mlp{}_p$ in Algorithm~\ref{alg:relnn}, respectively.
Note that, there is a different $\mlp_p$ for each predicate $p$.

\section{Learning the Value Function}

The architecture in Algorithm~\ref{alg:relnn} requires two inputs: a set of
atoms %
$\mathcal{A}$ and a set of objects %
$\mathcal{O}$.
Given a state $s$ over a set of objects $O$, along with a quantified goal $G$
over both a set of variables $V$ and the objects $O$, we need to transform these
into a single set of atoms, $\mathcal{A}$, and objects $\mathcal{O}$. The set of
objects $\mathcal{O} = O \cup V$ is set to contain the original object and the
variables, regarded as extra objects. The set of atoms $\mathcal{A}$ in turn, is
defined as
\begin{align*}
    \mathcal{A} = S & \cup \{P_G(\cdot) : P(\cdot) \in G\} \cup \{\textsc{Constant}(o) : o \in O\} \\
                    & \cup \{\textsc{Variable}(v) : v \in V\} \cup \mathcal{B}.
\end{align*}
Recall that for atoms in the goal, we use a specific goal predicate $P_G$,
instead of $P$, to extend the atoms in the
state~\cite{martin-geffner-ai2004}. In addition, we use the unary predicates
\textsc{Constant} and \textsc{Variable} to differentiate between true objects
and objects standing for %
variables. Finally, the set $\mathcal{B}$ contains a binary predicate, 
\textsc{PossibleBinding}, that enables communication between objects and variables. 
We define $\mathcal{B}$ as:
\[
    \mathcal{B} = \{\textsc{PossibleBinding}(o, v) : o \in O, v \in V\}.
\]
Without these atoms, constants, and variables cannot
communicate when the goal is fully quantified. This implies, for example, that
the final embedding of a variable does not depend on the current state.

The set of node embeddings at the last layer of the R-GNN is the result of the
net; that is, $\text{R-GNN}(\mathcal{A}, \mathcal{O}) = \{ f_L(o) \mid o \in
    \mathcal{O} \}$. Such embeddings are used to encode general value functions,
policies, or both. In this paper, we encode a learnable value function $V(S, G)$
through a simple additive readout that feeds the embeddings into an MLP:
\begin{displaymath}
    V(s, G) = \mlp{}\bigl(\textstyle\sum_{o \in \mathcal{O}} f_L(o)\bigr) \,.
\end{displaymath}

This value function will be used to iteratively guide a controller to ground
$G$, one variable at a time, until it is fully grounded. This means that $G$ can
be a partially quantified goal. Note that this is necessary because there is an
exponential number of possible bindings overall, but only a linear number of
possible bindings for a single variable. The loss we use for training
the R-GNN is the mean square error (MSE)
\begin{displaymath}
    \mathcal{L}(s, G) = \left(V(s, G) - V^*(s, G)\right)^2
\end{displaymath}
over the states $s$ and goals $G$ in the training set. The cost of a goal is
defined as the optimal number of steps required to achieve it from a given
state, denoted by $V^*(s, G)$. %
A state satisfies $G$ if there exists a complete binding such that the resulting
grounded goal is true in $s$. We compute $V^*(s, G)$ using breadth-first search
(BFS) from $s$ to determine the optimal distance to the nearest state that
satisfies $G$. If there is no reachable state that satisfies $G$, then we use a
large cost to indicate unsatisfiability (but not infinity). Generally, deciding
$V^*$ is computationally expensive; however, we train our networks on instances
with small state spaces.

\section{Expressivity Limitations}
\label{sec:expressivity}

In our experiments, we focus primarily on the use of a fixed set of colors.
This same set is also used in the example presented in Section~\ref{sec:example}.
Here, we explore why we chose to fix the set of colors and demonstrate that if the colors were allowed to vary as part of the input, R-GNNs would not be expressive enough.

It is important to determine if an object $o$ and a variable $x$ have the same color.
Generally, we represent a color with an object $c$ and express that an object $o$ has this color with the atom $\textsc{HasColor}(o, c)$.
Similarly, $\textsc{HasColor}(x, c)$ represents that a variable $x$ has color $c$.
To express that $o$ and $x$ share the same color, we use the formula:
\begin{align*}
    \textsc{SameColor}(o, x) =  \exists c :\; & \textsc{HasColor}(o, c) \\
                                              & \wedge \textsc{HasColor}(x, c).
\end{align*}
Since this formula uses three variables, it is likely not in $C_2$, meaning that R-GNNs cannot infer it.

Instead, we use a fixed set of unary predicates $\textsc{C}_1, \dots, \textsc{C}_n$ to denote colors.
This allows us to express the \textsc{SameColor} relation as:
\begin{align*}
    \textsc{SameColor}(o, x) = & \left[\textsc{C}_1(o) \wedge \textsc{C}_1(x)\right] \vee \dots \\
                               & \vee \left[\textsc{C}_n(o) \wedge \textsc{C}_n(x)\right].
\end{align*}
With only two variables used, this approach allows R-GNNs to potentially learn whether two objects or variables share the same color, as long as the number of colors is fixed.
For clarification, we do not use the \textsc{SameColor} predicate explicitly in our experiments, since it can be inferred using \textsc{PossibleBinding} along with a fixed set of colors.

In conclusion, we use a fixed set of colors due to the expressivity limitations of the R-GNN architecture, where the color predicates are part of the domain.
We also highlight that not every unique variable in the quantified goal contributes to the $C_2$ limit.
This is because some variables do not overlap in scope, allowing them to be reused, which helps keep the total number of variables down.
This is illustrated in Section~\ref{sec:analysis}, where many variables are used in the goal, but their total number is bounded by a fixed constant.

\section{Domains}

Now, we will describe the domains used in our experiments and the types of goals
that the architecture must learn to ground. All domains are taken from the
International Planning Competition (IPC), but are augmented with colors. With
colors, we can express a richer class of goals, where objects can be
referred to by their (non-unique) colors. However, the overall approach is
general, and descriptions involving other object attributes such as size or
shape and their combinations can also be used,  
if such predicates are used in the training instances.

\subsection{Blocks}

\begin{figure}[t]
    \centering
    \subfigure[Grounded state.]{%
        \includegraphics[scale=0.33]{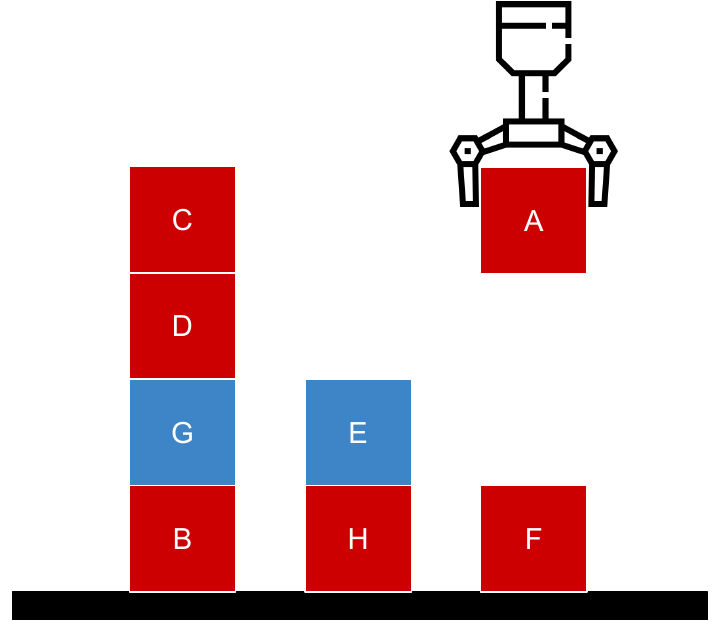}
        \label{fig:blocks-example}
    }
    \subfigure[Quantified goal.]{%
        \includegraphics[scale=0.33]{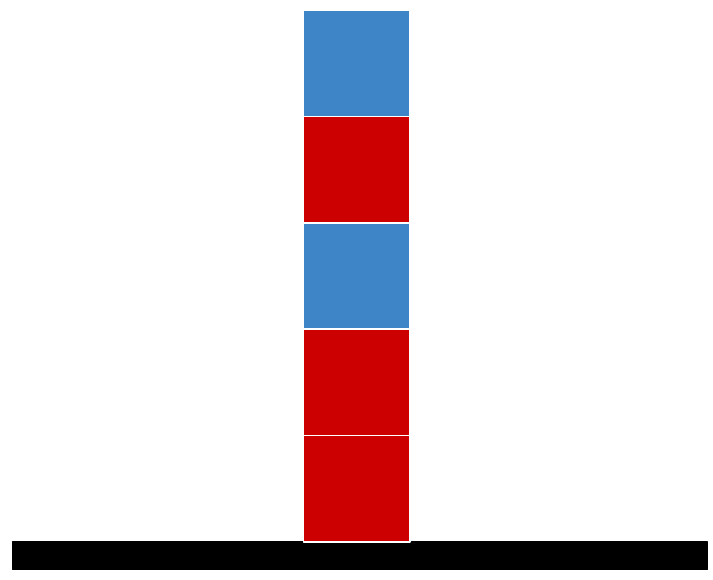}
        \label{fig:blocks-goal-example}
    } \caption{%
        A Blocks example. %
        The goal shown on the right is: $ \exists x_1, \dots x_5 : \textsc{Blue}(x_1) \wedge
        \textsc{Red}(x_2) \wedge \textsc{Blue}(x_3) \wedge \textsc{Red}(x_4)
        \wedge \textsc{Red}(x_5) \wedge \textsc{On}(x_1, x_2) \wedge
        \textsc{On}(x_2, x_3) \wedge \textsc{On}(x_3, x_4) \wedge
        \textsc{On}(x_4, x_5)$. The cost of an optimal goal grounding is 9. In
        the experiments, our learned model instead finds a suboptimal goal
        grounding with a cost of 15.}
    \label{fig:blocks}
\end{figure}

Blocks is the standard Blockworld domain with a gripper and block colors
that can be used in the goals. Figure~\ref{fig:blocks} depicts an instance of
the domain involving 8 blocks, named from $A$ to $H$, 
and expressed as:

\begin{align*}
    \exists x_1, \ldots, x_5 :\; & \textsc{Blue}(x_1) \wedge \textsc{Red}(x_2) \wedge \textsc{Blue}(x_3) \\
                                 & \wedge \textsc{Red}(x_4) \wedge \textsc{Red}(x_5) \wedge \textsc{On}(x_1, x_2) \\
                                 & \wedge \textsc{On}(x_2, x_3) \wedge \textsc{On}(x_3, x_4) \wedge \textsc{On}(x_4, x_5).
    \label{eq:blocks-goal}
\end{align*}

This goal requires constructing a tower in which a blue block $x_1$ rests on a
red block $x_2$, which in turn rests on a blue block $x_3$, which rests on a red
block $x_4$, and finally, that rests on another red block $x_5$. The optimal
grounding involves binding block $x_5$ to block $F$ and block $x_4$ to block
$A$, so that the target tower is built on top of $F$.

Let us emphasize that the grounding problem tackled by the learner is not easy.
Firstly, the resulting binding needs to be \emph{logically valid}, meaning it
must adhere to the static constraints of the goal, such as colors in this case.
A binding would be logically invalid if a variable $x_i$ is bound to a constant
$c$ with a different color from $x_i$ in the goal. Secondly, the resulting
binding must lead to a \emph{reachable goal}. In this domain, it implies that no
pair of variables $x_i$ can be bound to the same constant, as the blocks $x_i$
need to form a single tower. Lastly, the resulting binding needs to be optimal;
that is, it must be achievable in the fewest steps possible. None of these three
properties -- \emph{validity}, \emph{reachability}, or \emph{optimality} -- are
hardcoded or guaranteed by the learning architecture, but we can experimentally
test them. Planners that handle existentially quantified goals by expanding them
into grounded DNF formulas ensure and exploit validity. By pruning the DNF terms
that are not logically valid, they exploit (but do not guarantee) reachability
by removing DNF clauses containing mutually exclusive atom pairs.%
\footnote{They cannot precompute all mutually exclusive pairs. Additionally,
    states may remain unreachable even if they do not include a mutually exclusive
    pair of atoms. For instance, in blocks, states where a block is above itself,
    but not directly on itself, will be unreachable, despite lacking any mutually
    exclusive pairs.}
They ensure optimality if they are optimal planners. In our setting, validity,
reachability, and optimality, or a reasonable approximation, must be learned
solely from training.

The domain called Blocks-C is essentially the same as the Blocks domain, but
instead of focusing on the relation \textsc{On}, it revolves around the relation
\textsc{Clear}. This implies that the goal is to dismantle towers to
uncover blocks of specific colors, with the fewest steps possible.

\subsection{Gripper}

\begin{figure}[t]
    \centering
    \subfigure[Grounded state.]{
        \includegraphics[scale=0.33]{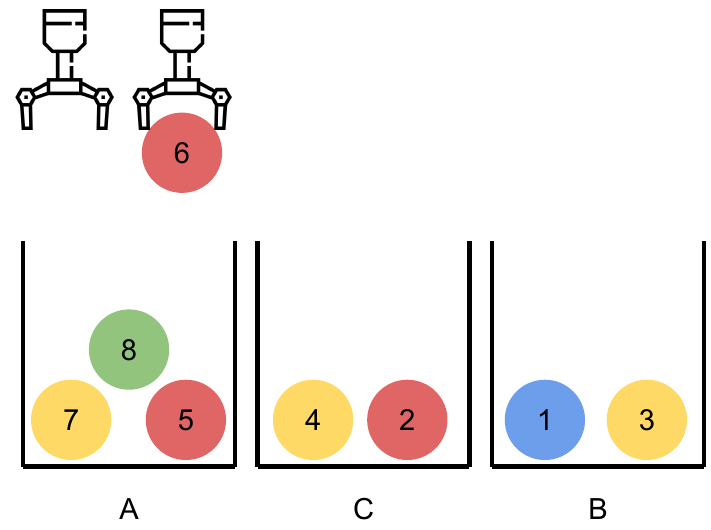}
        \label{fig:gripper-example-state}
    }
    \subfigure[Predicted grounding.]{
        \includegraphics[scale=0.33]{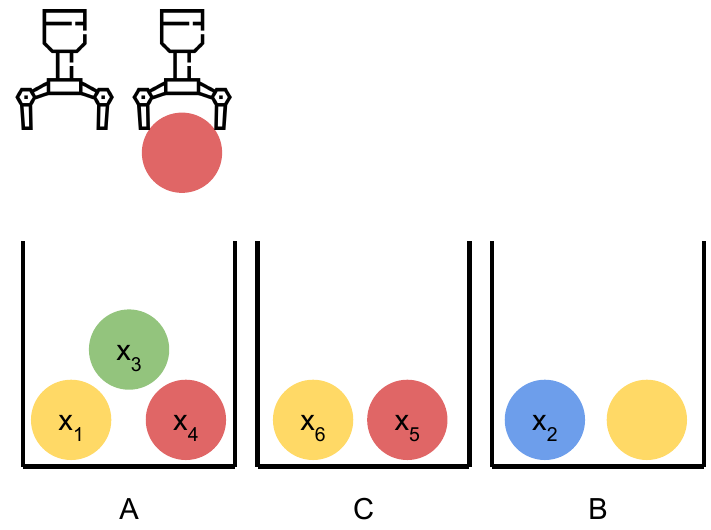}
        \label{fig:gripper-example-goal}
    }

    \caption{A Gripper example. %
      The goal is: $ \exists x_1, \ldots, x_4 : \big[ \textsc{Yellow}(x_1)
        \wedge \textsc{Blue}(x_2) \wedge \textsc{Green}(x_3) \wedge
        \textsc{Red}(x_4) \wedge \textsc{Red}(x_5) \wedge \textsc{Yellow}(x_6)
        \wedge \textsc{At}(x_1, A) \wedge \textsc{At}(x_2, A) \wedge
        \textsc{At}(x_3, B) \wedge \textsc{At}(x_4, A) \wedge \textsc{At}(x_5,
        C) \wedge \textsc{At}(x_6, C) \big] \wedge \big[ \bigwedge_{i,j}
        \textsc{Neq}(x_i,x_j) \big]$, where $\textsc{Neq}(x, y)$ denotes that
        $x$ and $y$ have to be different constants. The grounding of the goal
        that our learned model predicts is shown on the right. This is an
        optimal binding with a cost of 8.}
    \label{fig:gripper-example}
\end{figure}

In Gripper, there is a robot equipped with two grippers that can pick up balls
and move them between rooms. In our version, the balls are colored, and there is
a fixed number of rooms that are not necessarily adjacent to each other.
Figure~\ref{fig:gripper-example} provides an example. In this figure, there are
three rooms labeled as $A$, $B$, and $C$. The goal is: %
\begin{align*}
    \exists x_1, \ldots, x_4 :\; & \textsc{Yellow}(x_1) \wedge \textsc{Blue}(x_2) \wedge \textsc{Green}(x_3) \\
                                 & \wedge \textsc{Red}(x_4) \wedge \textsc{Red}(x_5) \wedge \textsc{Yellow}(x_6) \\
                                 & \wedge \textsc{At}(x_1, A) \wedge \textsc{At}(x_2, A) \wedge \textsc{At}(x_3, B) \\
                                 & \wedge \textsc{At}(x_4, A) \wedge \textsc{At}(x_5, C) \wedge \textsc{At}(x_6, C) \\
                                 & \wedge \big[ \bigwedge_{i,j} \textsc{Neq}(x_i,x_j) \big]
\end{align*}
The goal is to place a yellow ball $x_1$, a blue ball
$x_2$ and a red ball $x_4$ in room A, a green ball $x_3$ in room B, and a red ball $x_5$ and a yellow ball $x_6$ in room C.
The optimal solution is to bind $x_1$ to $7$, $x_4$ to $5$, $x_5$ to $2$ and $x_6$ to $4$, since then the
goal atoms with these variables are already true in the initial state.
As for $x_2$ and $x_3$, these need to be bound to $1$ and $8$ respectively,
since these are the only balls of the correct color.

\subsection{Delivery}

\begin{figure}[t]
    \centering
    \subfigure[Grounded state.]{%
        \includegraphics[scale=0.32]{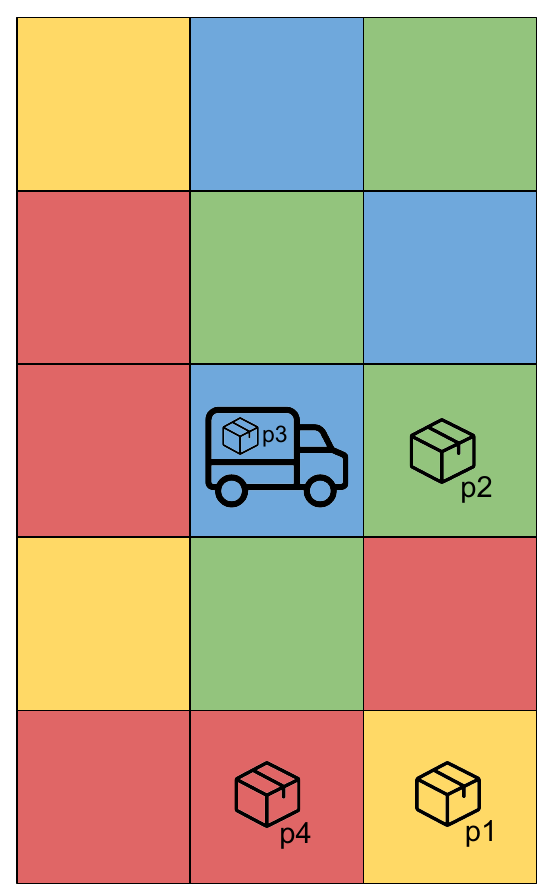}
        \label{fig:delivery-example-state}
    }
    \subfigure[Predicted grounding.]{
        \includegraphics[scale=0.32]{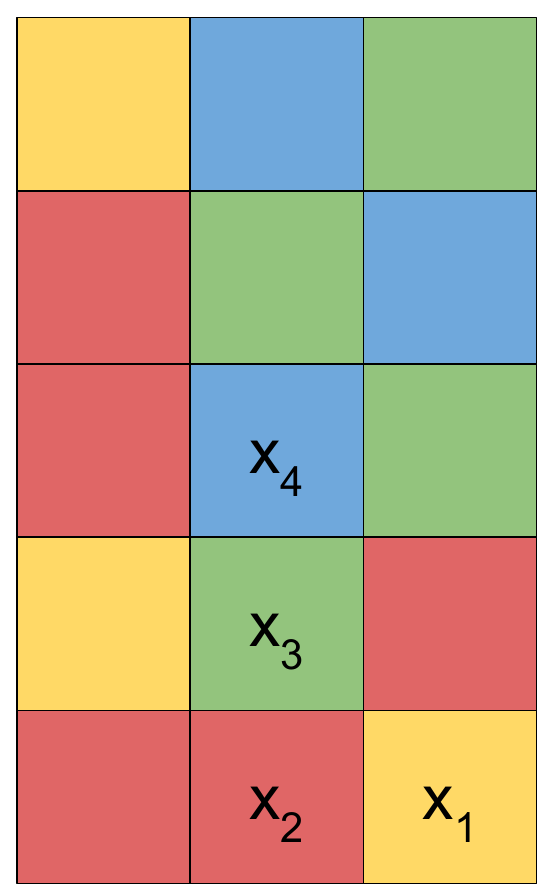}
        \label{fig:delivery-example-goal}
    }
    \caption{A Delivery example. %
          The goal is:
        $\exists x_1, \dots, x_4 : \textsc{At}(p_1, x_3) \wedge \textsc{At}(p_3,
            x_2) \wedge \textsc{At}(p_4, x_1) \wedge \textsc{At}(t, x_4) \wedge
            \textsc{Yellow}(x_1) \wedge \textsc{Red}(x_2) \wedge \textsc{Green}(x_3)
            \wedge \textsc{Blue}(x_4) \wedge \big[ \bigwedge_{i,j} \textsc{Neq}(x_i,
            x_j) \big]$, where $\textsc{Neq}(x, y)$ denotes that $x$ and $y$ have to be
            different constants.
            Figure \ref{fig:delivery-example-goal} shows the bindings of the learned model.
            This binding has a cost of 11, while LAMA produces a plan with a cost of 17.}
    \label{fig:delivery-example}
\end{figure}

In this domain, there is a truck, several packages, and a %
grid with %
colored cells. The goal is to distribute the packages to cells with
specific colors. In the following example, there are $3$ packages, $p_1$, $p_2$,
and $p_3$, to be distributed using truck $t$:
\begin{align*}
    \exists x_1, \dots, x_4 :\; & \textsc{At}(p_1, x_3) \wedge \textsc{At}(p_3, x_2) \wedge \textsc{At}(p_4, x_1) \\
                                & \wedge \textsc{At}(t, x_4) \wedge \textsc{Yellow}(x_1) \wedge \textsc{Red}(x_2)  \\
                                & \wedge \textsc{Green}(x_3) \wedge \textsc{Blue}(x_4) \\
                                & \wedge \big[ \bigwedge_{i,j} \textsc{Neq}(x_i, x_j) \big]
\end{align*}
To distribute these packages efficiently, the truck must decide on an order to pick up 
the packages and deliver them to cells close to this planned route to
minimize the total cost.

In this example, the truck must deliver package $p_1$ to a green cell $x_3$,
$p_3$ to a red cell $x_2$, $p_4$ to a yellow cell $x_1$, and then park the truck
$t$ in a blue cell $x_4$. As %
seen in Figure \ref{fig:delivery-example-goal},
the learned model selects cells close to where the packages are, which lowers the
total distance traveled by the truck.

\subsection{Visitall}

\begin{figure}[t]
    \centering
        \includegraphics[scale=0.18]{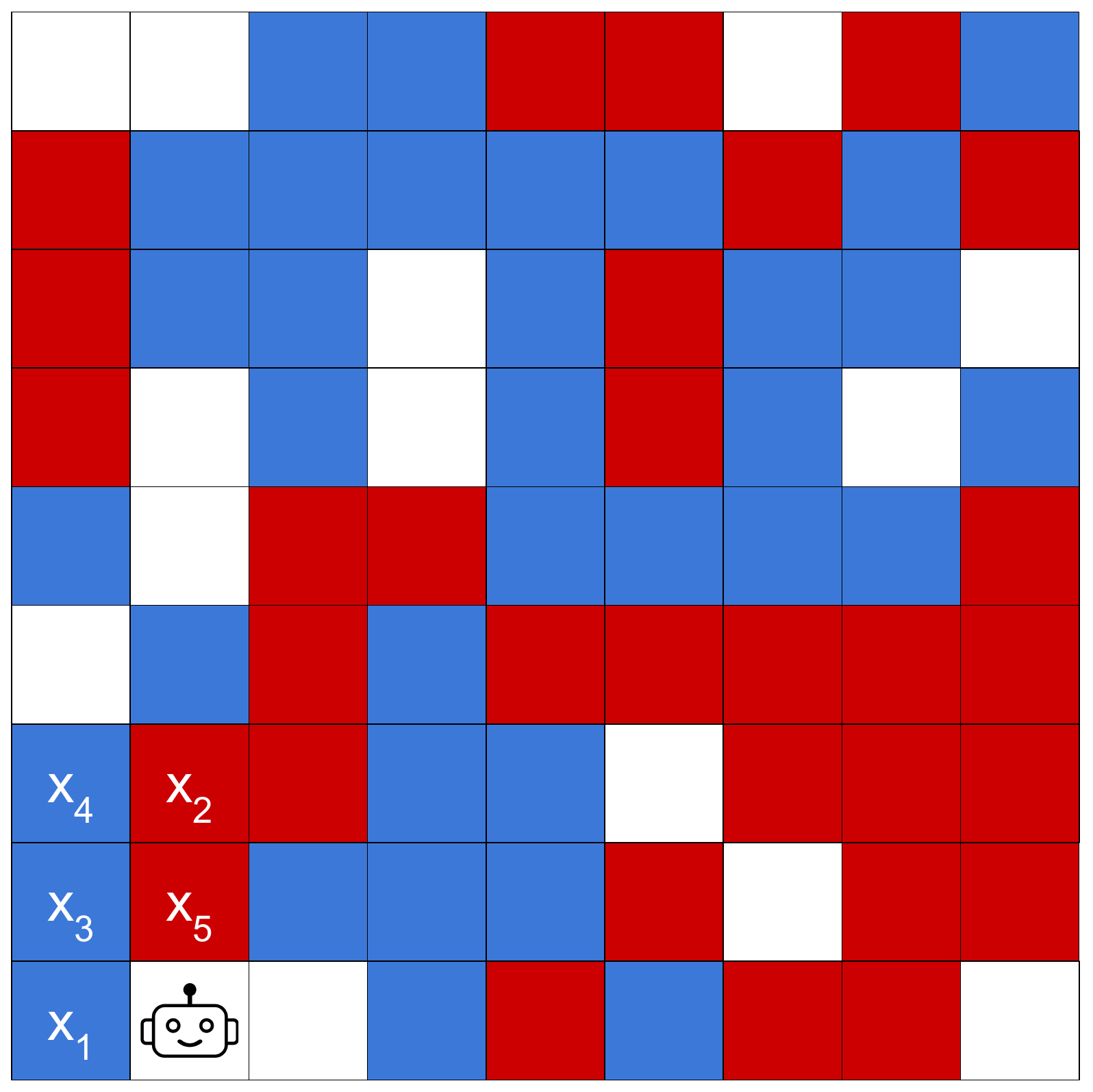}
    \caption{%
        A Visitall example. %
        The goal is: $\exists
        x_1, \dots, x_5 : \big[ \textsc{Blue}(x_1) \wedge \textsc{Blue}(x_3)
        \wedge \textsc{Blue}(x_4) \wedge \textsc{Red}(x_2 ) \wedge
        \textsc{Red}(x_5) \wedge \textsc{Visited}(x_1) \wedge
        \textsc{Visited}(x_2) \wedge \textsc{Visited}(x_3) \wedge
        \textsc{Visited}(x_4) \wedge \textsc{Visited}(x_5) \big] \wedge \big[
        \bigwedge_{i,j} \textsc{Neq}(x_i, x_j) \big]$, where $\textsc{Neq}(x,
        y)$ denotes that $x$ and $y$ have to be different constants. An optimal
        grounding with a cost of $5$ is shown. In this case,  %
        LAMA fails to find a solution within the time limit, but
        when combined with our learned model, it solves the resulting grounded instance optimally.}
    \label{fig:visitall-example}
\end{figure}

In a more general version of Visit-1, presented in Section~\ref{sec:example},
the robot is required to visit at least one cell of each color specified in the
goal while also minimizing the total distance traveled. Figure~\ref{fig:visitall-example} provides an example. 
The example goal is: %
\begin{align*}
    \exists x_1, \dots, x_5 : \big[ & \textsc{Blue}(x_1) \wedge \textsc{Blue}(x_3) \wedge \textsc{Blue}(x_4)          \\
                                    & \wedge \textsc{Red}(x_2 ) \wedge \textsc{Red}(x_5) \wedge \textsc{Visited}(x_1) \\
                                    & \wedge \textsc{Visited}(x_2) \wedge \textsc{Visited}(x_3)                       \\
                                    & \wedge \textsc{Visited}(x_4) \wedge \textsc{Visited}(x_5) \big]                 \\
                                    & \wedge \big[ \bigwedge_{i,j} \textsc{Neq}(x_i, x_j) \big]
\end{align*}
The robot needs to visit 3 blue cells $ x_1 $, $ x_3 $, and $ x_4 $, and 2 red
cells $ x_2 $ and $ x_5 $. An optimal solution with a cost of $5$ is shown in
Figure~\ref{fig:visitall-example}. This domain poses a difficult optimization
problem, similar to but distinct from the Traveling Salesman Problem (TSP).

\begin{table*}[ht]
    \centering
    \footnotesize
    \begin{tabular}{lr|rrr|rrr|rrr}
        \hline
                 & \multicolumn{1}{c}{} & \multicolumn{3}{c}{No $x_i \neq x_j$ constraint} & \multicolumn{3}{c}{With $x_i \neq x_j$ constraints} & \multicolumn{3}{c}{Random groundings}                                                                    \\
        Domain   & \#                   & Cov.                                             & $V^*$                                               & $V / V^*$                             & Cov.    & $V^*$ & $V / V^*$ & All Cov. & Valid Cov. & $ V / V^*$ \\ \hline
        Blocks   & 500                  & 99.8 \%                                          & 8.918                                               & 1.211                                 & 99.8 \% & 8.918 & 1.211     & 3.8  \%  & 76.2 \%    & 1.391      \\
        Blocks-C & 500                  & 99.8 \%                                          & 3.402                                               & 1.103                                 & 100  \% & 4.478 & 1.019     & 6.8  \%  & 100  \%    & 1.797      \\
        Gripper  & 500                  & 100  \%                                          & 4.78                                                & 1.298                                 & 99.4 \% & 5.304 & 1.189     & 3    \%  & 86.6 \%    & 1.499      \\
        Delivery & 500                  & 99.8 \%                                          & 9.016                                               & 1.238                                 & 100  \% & 9.224 & 1.177     & 10.8 \%  & 100  \%    & 1.522      \\
        Visitall & 500                  & 100  \%                                          & 4.072                                               & 1.181                                 & 99.2 \% & 4.712 & 1.080     & 2.6  \%  & 100  \%    & 1.445      \\
        \hline
    \end{tabular}
    \caption{%
        This table shows how well the learned model performed on the \textbf{optimal test sets} (refer to Table~\ref{tab:domains}).
        For each domain, we evaluate the learned models in two settings: one where multiple variables assigned to the same object can result in valid bindings, 
        and one where these are always invalid.
        These settings are labeled as "No $x_i \neq x_j$ constraint" and "With $x_i \neq x_j$ constraints," respectively.
        The "\#" column indicates the number of test instances.
        The performance of the learned models is displayed as coverage (Cov.), which represents the percentage of instances solved after grounding the goal.
        The average cost of optimal plans is shown in the $V^*$ column, and the quality of the grounded goal is presented as the ratio 
        between the cost of the plan found using the grounded goal and the cost of an optimal plan in the $V / V^*$ column.
        Additionally, we compare the results to two baselines where groundings are randomly sampled: 
        one considering all possible bindings and another considering only those that satisfy static constraints (e.g., colors).
        These baselines are labeled as "Random groundings," with "All" and "Valid" denoting the two methods, respectively.
        The quality ratio presented in $V / V^*$ corresponds to the "Valid" baseline.
    }
    \label{tab:performance-opt}
\end{table*}

\begin{table}[ht]
  \centering
  \footnotesize
  \begin{tabular}{lrrr}
    \hline
             &                           & \multicolumn{1}{c}{Optimal} & \multicolumn{1}{c}{LAMA} \\
    Domain   & \multicolumn{1}{c}{Train} & \multicolumn{1}{c}{Test}    & \multicolumn{1}{c}{Test} \\\hline
    Blocks   & [2, 7]                    & 8                           & [9, 17]                  \\
    Blocks-C & [2, 7]                    & 8                           & [9, 17]                  \\
    Gripper  & [9, 11]                   & 13                          & [15,47]                  \\
    Delivery & [6, 25]                   & [26, 30]                    & [31, 88]                 \\
    Visitall & [2, 16]                   & 20                          & [25, 100]                \\ \hline
  \end{tabular}
  \caption{%
    An overview of the different datasets %
    used in our experiments.
    The intervals represent the number of constants appearing in the states. For
    instance, we train a model using up to $7$ blocks for the Blocks domain.
    Subsequently, we evaluate this model on two different test sets: one where
    we were able to compute the cost of an optimal binding, labeled as
    \emph{Optimal}, and another where the cost is determined using the LAMA
    planner, labeled as \emph{LAMA}. In all domains, the training set
    includes instances with $1$ to $4$ goal variables, and the test data includes
    instances with $1$ to $6$ goal variables. The number of colors for each instance
    ranges from $1$ to $6$ for both the training and test sets.}
  \label{tab:domains}
\end{table}

\begin{table*}[ht]
    \centering
    \footnotesize
    \begin{tabular}{lr|rrrrr|rrrrr}
        \hline
                 & \multicolumn{1}{c}{} & \multicolumn{5}{c}{No $x_i \neq x_j$ constraint} & \multicolumn{5}{c}{With $x_i \neq x_j$ constraints}                                                                                   \\
        Domain   & \#                   & Cov.                                             & LAMA Cov.                                           & $V^L$ & $V / V^L$ & Speedup & Cov.    & LAMA Cov. & $V^L$ & $V / V^L$ & Speedup \\ \hline
        Blocks   & 500                  & 100  \%                                          & 99.4 \%                                             & 11.93 & 1.245     & 19.837  & 100  \% & 99.2 \%   & 11.93 & 1.245     & 19.837  \\
        Blocks-C & 500                  & 100  \%                                          & 100  \%                                             & 2.74  & 1.179     & 1.359   & 99.8 \% & 100  \%   & 4.438 & 1.047     & 1.190   \\
        Gripper  & 500                  & 97.6 \%                                          & 96.2 \%                                             & 6.368 & 1.263     & 104.335 & 99.6 \% & 97   \%   & 7.07  & 1.239     & 68.306  \\
        Delivery & 500                  & 99   \%                                          & 98.6 \%                                             & 7.958 & 1.459     & 50.495  & 100  \% & 98.6 \%   & 8.768 & 1.212     & 36.533  \\
        Visitall & 500                  & 100  \%                                          & 98   \%                                             & 4.884 & 1.480     & 33.219  & 88.8 \% & 96.6 \%   & 5.306 & 1.122     & 43.468  \\
        \hline
    \end{tabular}
    \caption{%
        This table shows the performance of the learned model on the \textbf{LAMA test sets} (refer to Table~\ref{tab:domains}).
        See Table~\ref{tab:performance-opt} for explanations of most columns.
        We were unable to compute $V^*$ for these instances, so we approximate it using LAMA, denoted as $V^L$.
        We evaluate two configurations: first, LAMA using grounded goals obtained through our learned models, and second, LAMA using the quantified goal.
        The coverage of the first and second configurations is presented in the "Cov." and "LAMA Cov." columns, respectively.
        The "Speedup" column illustrates the speedup of using the first configuration over the second.
    }
    \label{tab:performance-lama}
\end{table*}

\section{Experimental Results}
\label{sec:experiments}

Once a model has been trained to predict the cost of partially quantified goals
relative to a state, it is then possible to use it as a \emph{policy}. If $V$ is
a learned model for some domain, then it defines a policy $\pi_V$ over the
\emph{space of partially quantified goals} as follows. Let $s$ and $G$ be a
state and a %
goal, respectively; the successors $N(s, G)$ of
$s$ and $G$ are a set of
goals where $G$ has bound a single
variable with a constant. The policy $\pi_V$ is:
\[
  \pi_V(s, G) = \argmin_{G_{x=c} \in N(s, G)} V(s, G_{x=c}).
\]
The policy is iteratively used until the goal is fully grounded. By binding a
single variable, we avoid the combinatorial explosion that would occur when
considering all possible combinations. The number of iterations required to
fully ground $G$ is equal to the number of variables.

We implemented the proposed method using PyTorch\footnote{Code, data, and models:
\url{https://zenodo.org/records/13235160}} and trained the models on NVIDIA A10 GPUs
with 24 GB of memory. Training lasted a maximum of $100~000$ epochs or $72$
hours. We used Adam~\cite{kingma-et-al-iclr2015} with a learning rate of
$0.001$, using batch sizes ranging from $1024$ to $8192$. Each domain's dataset
comprised $40~000$ pairs of states and goals, $39~500$ for
training and 500 for validation. The number of constants used in both the
training and testing sets are detailed in Table~\ref{tab:domains}. During
training, we used up to $4$ variables, while during testing, we used up to $6$.
Furthermore, the trained models support up to $6$ distinct colors. The learned
models have $30$ layers and an embedding size of $32$. For each specific domain,
we trained a single model and the model with the lowest validation loss was used for
testing.

\subsection{Results}

As mentioned earlier, our goal is to learn goal groundings that are:
\emph{logically valid}, \emph{reachable}, and \emph{efficient} (close to optimal).
Tables~\ref{tab:performance-opt} and~\ref{tab:performance-lama} show the
performance of our learned models across various domains.
Tables~\ref{tab:scaling-blocks-on} and~\ref{tab:scaling-visitall} delve deeper
into two specific domains, Blocks and Visitall, illustrating how well the
learned models scale and generalize with an increasing number of constants.
We discuss how the experiments shed light on these properties.

\medskip \noindent \textbf{Validity and reachabilty.} %
A valid grounding is one that satisfies static constraints, such as colors. For
instance, the learned models are permitted to bind a variable to a red block
even if that variable must be bound to a blue block. If this occurs, the
resulting grounded goal is unsatisfiable, leading to an unsolvable instance. In
Tables~\ref{tab:performance-opt} and~\ref{tab:performance-lama}, the coverage
measures this property, as all solvable groundings must also be valid. However,
a valid grounding does not necessarily have to be reachable from the given
state. The coverage measures that the goal groundings are both valid and
reachable. In our experiments, our learned models result in a system with very
high coverage, meaning that the vast majority of the goal groundings are both
valid and reachable. Additionally, in Table~\ref{tab:performance-opt}, the
baseline "All Cov." shows that randomly grounding variables will likely
result in an invalid grounding. The other baseline, "Valid Cov.", randomly
selects a valid grounding, ensuring validity but not reachability. However, it
often leaves Blocks and Gripper instances unsolvable. In contrast, our learned
models produce goal groundings that are both valid and reachable.

\medskip \noindent \textbf{Efficiency.} %
We evaluate efficiency in two datasets. Firstly, in
Table~\ref{tab:performance-opt}, we compare the actual cost of predicted goal
groundings, denoted as $V$, to the optimal cost of goal groundings, denoted as
$V^*$. Secondly, in Table~\ref{tab:performance-lama}, where determining $V^*$ is
computationally infeasible, we instead compare against the plan length
identified by LAMA~\cite{richter-westphal-jair2010}. These comparisons are presented as the quotients
$V/V^*$ and $V/V^L$ in Tables~\ref{tab:performance-opt} and
\ref{tab:performance-lama}, respectively. We observe that our goal groundings
are often close to optimal for smaller instances, with averages reaching up to
$1.238$. For larger instances requiring LAMA, goal groundings degrade a bit,
with averages reaching up to $1.48$. It is important to note that these ratios
are calculated only over instances solvable by both LAMA alone and LAMA with our
learned models and that LAMA does not necessarily find an optimal plan for
grounded goals. Tables~\ref{tab:scaling-blocks-on} and
\ref{tab:scaling-visitall} show whether the quality of the goal groundings
deteriorates as the instance size increases in two domains: Blocks and Visitall.
It seems that the quality does not degrade, as it remains consistent when the
number of constants increases.

\medskip \noindent \textbf{Scalability.} %
LAMA compiles quantified goals by expanding them into grounded DNF formulas,
whose clauses become the conditions of axioms that derive a new dummy atom. This
expansion is exponential in the number of goal variables, takes preprocessing
time, and affects the quality of the resulting heuristic. As a result, if LAMA
is given a high-quality \emph{grounded} goal, it can avoid preprocessing and get
a more informed heuristic. The column labeled "Speedup" in
Tables~\ref{tab:performance-opt} and~\ref{tab:performance-lama} indicates the
speedup achieved by running LAMA with our learned models compared to LAMA alone.
We observe a two-order-of-magnitude speedup in the Gripper domain in
Table~\ref{tab:performance-lama}, and in most domains, we see a
one-order-of-magnitude speedup. Furthermore, in
Tables~\ref{tab:scaling-blocks-on} and~\ref{tab:scaling-visitall}, instances are
aggregated based on size (\# constants) rather than by domain. This allows us to
observe scalability in the Blocks and Visitall domains, demonstrating that the
speedup increases with the instance size. In Table~\ref{tab:scaling-visitall},
as the number of objects becomes very large, the coverage decreases for both
LAMA alone and LAMA with the learned model: in the first case, the search times
out; in the second, unreachable grounded goals are generated.

\begin{table}[tp]
    \centering
    \footnotesize
    \begin{tabular}{lrrrrrr}
        \hline
        Size & \# & Cov.   & L. Cov. & $V^L$  & $V / V^L$ & Speedup \\ \hline
        7    & 33 & 100 \% & 100  \% & 18.485 & 1.131     & 0.453   \\
        9    & 33 & 100 \% & 100  \% & 15.939 & 1.399     & 4.207   \\
        11   & 31 & 100 \% & 100  \% & 14.645 & 1.198     & 16.818  \\
        13   & 27 & 100 \% & 100  \% & 14.111 & 1.491     & 1.843   \\
        15   & 33 & 100 \% & 100  \% & 11.545 & 1.247     & 6.964   \\
        17   & 29 & 100 \% & 93.1 \% & 11.778 & 1.476     & 136.028 \\ \hline
    \end{tabular}
    \caption{%
        Performance and generalization scaling within the domain of \textbf{Blocks}.
        The instances are scaled by the number of constants (blocks), as shown in the column labeled "Size".
        Refer to Table~\ref{tab:performance-lama} for explanations of the other columns.
        Note that, the largest instance in the training set consists of exactly 7 constants.
        Therefore, this table also illustrates how well the learned models generalize beyond the training distribution.
    }
    \label{tab:scaling-blocks-on}
\end{table}

\begin{table}[ht]
    \centering
    \footnotesize
    \begin{tabular}{lrrrrrr}
        \hline
        Size & \# & Cov. & L. Cov. & $V^L$ & $V/V^L$ & Speedup \\ \hline
        16   & 27 & 96.3 & 100     & 5.444 & 1.003   & 0.788   \\
        20   & 25 & 100  & 100     & 6.36  & 1.126   & 1.775   \\
        25   & 25 & 96   & 100     & 6.0   & 1.132   & 5.333   \\
        49   & 21 & 71.4 & 95.2    & 8.2   & 0.911   & 13.813  \\
        64   & 32 & 56.3 & 81.3    & 7.577 & 1.027   & 14.996  \\
        100  & 36 & 61.1 & 80.6    & 4.690 & 1.386   & 8.513   \\ \hline
    \end{tabular}
    \caption{%
        Performance and generalization scaling within the domain of \textbf{Visitall}.
        See Table~\ref{tab:scaling-blocks-on} for explanations of the columns.
    }
    \label{tab:scaling-visitall}
\end{table}

\section{Analysis: Visit-Many}
\label{sec:analysis}

We now continue the example from Section~\ref{sec:example} and present a detailed analysis where a specific number of colors must be visited.
In our experiments, we allow the number of colors in the goal to vary, but here we assume exactly $k$ colors:
\[
    \exists x_1, \dots, x_k : \textsc{C}_1(x_1) \wedge \dots \wedge \textsc{C}_k(x_k), \text{where } \textsc{C}_i \in \mathcal{C}.
\]

To determine distances, we adjust the Boolean functions $\text{P}$ and $\text{SP}$ to check if there is a path of length $d$ that visits cells with colors $\langle \textsc{C}_1, \dots, \textsc{C}_n \rangle$ in the specified order:
\begin{align*}
    \text{P}_{0, C}(x) &= \textsc{C}(x), \\
    \text{P}_{d, \langle \textsc{C}_1, \dots, \textsc{C}_n \rangle}(x) \hspace{-1pt}&=\hspace{-1pt} \textsc{C}_1(x) \wedge P_{d, \textsc{C}_2, \dots, \textsc{C}_n}(x), \\
    \text{P}_{d, \langle \textsc{C}_1, \dots, \textsc{C}_n \rangle}(x) \hspace{-1pt}&=\hspace{-1pt} \exists y : \textsc{Conn.}(x, y) \wedge P_{d - 1, \langle \textsc{C}_1, \dots, \textsc{C}_n \rangle}(y), \\
    \text{SP}_{d, \langle \textsc{C}_1, \dots, \textsc{C}_n \rangle}(x) \hspace{-1pt}&=\hspace{-1pt} \text{P}_{d, \langle \textsc{C}_1, \dots, \textsc{C}_n \rangle}(x) \wedge \neg \text{P}_{d - 1, \langle \textsc{C}_1, \dots, \textsc{C}_n \rangle}(x).
\end{align*}
The colors in the subscript act like a queue: if the color of the current cell matches the color in front of the queue, that color is removed.
The value function is then defined as:
\begin{align*}
    \textsc{G}_{\bar{\textsc{C}}} &= \neg \exists \textsc{C} \in \mathcal{C} \setminus \bar{\textsc{C}} : \left[ \exists v : \textsc{C}_G(v) \right], \\
    \text{D}_{d, \bar{\textsc{C}}} &= \textsc{G}_{\bar{\textsc{C}}} \wedge \exists x : \left[ \textsc{At-Robot}(x) \wedge \text{SP}_{d, \bar{\textsc{C}}}(x) \right], \\
    V^*(s; G) &= \min_{d \in \mathcal{D}, \bar{\textsc{C}} \in \mathcal{C}^k} d \cdot \bracket{\text{D}_{d, \bar{\textsc{C}}}} + N \cdot \bracket{\neg D_{d, \bar{\textsc{C}}}}.
\end{align*}
Here, $\mathcal{D}$ is the fixed set of distances, $\mathcal{C}$ is the fixed set of all colors, and $\mathcal{C}^k$ is the set of all ordered sequences of length $k$, where each element is chosen from $\mathcal{C}$ (elements can be repeated).
The $\textsc{G}_{\bar{\textsc{C}}}$ feature is false when the goal involves a color not in the sequence $\bar{\textsc{C}}$.
This is because both the goal and the sequence allow for repetition.
Since the number of colors is fixed, we can expand the outer quantifier.
We assume that the given state does not contain any $\textsc{Visited}$ atoms, since grounding should be done on the initial state.
If there is such an atom, $V^*$ may return a suboptimal value.

No more than two variables are used at any point, indicating that this domain with such quantified goals belongs to $C_2$.
Note that this analysis excludes any inequality constraints, which might move the task outside of $C_2$.

\section{Conclusions}

We considered the problem of learning to ground existentially quantified goals, building on existing techniques for learning general policies using GNNs.
These methods usually require goals that are already grounded, and the aim is to enable them to support existentially quantified goals by grounding them beforehand.
This preprocessing step is also useful for classical planning, where existentially quantified goals exponentially increase time and space requirements.
We also discuss this approach in relation to $C_2$ logics, as it builds on relational GNNs for representing and learning general value functions~\cite{stahlberg-et-al-icaps2022}.

Grounding existentially quantified goals after learning is computationally efficient and does not require search.
The process involves greedily following the learned value function, substituting one variable at a time with a constant.
However, learning such a value function is challenging because the grounded goals must satisfy conditions of logical validity, reachability, and optimality.
We tested these conditions experimentally on instances with more objects, goal atoms, and variables.
In the future, we plan to explore unsupervised methods for grounding learning, fine-tuning weights using reinforcement learning as done by~\cite{stahlberg-et-al-kr2023}, and experimenting with GNN architectures with improved expressive power~\cite{barcelo-et-al-iclr2020,grohe-lics2021}.
This might include adding memory and recursion capabilities~\cite{pfluger-et-al-aaai2024}.

\section*{Acknowledgements}

This work has been supported by the Alexander von Humboldt Foundation with funds from the Federal Ministry for
Education and Research. It has also received funding from the European Research Council (ERC), Grant agreement No
885107, the Excellence Strategy of the Federal Government and the NRW Lander, Germany,
and the Knut and Alice Wallenberg (KAW) Foundation under the WASP program.
The computations were enabled by resources provided by the National Academic Infrastructure for Supercomputing in Sweden (NAISS),
partially funded by the Swedish Research Council through grant agreement no. 2022-06725.

\bibliographystyle{kr}
\bibliography{bibliography,crossref}

\end{document}